# Assessing socio-economic climate impacts from text data

Mariana Madruga de Brito[1*], Brielen Madureira[2,1], Taís Maria Nunes Carvalho[1,2], Damien Delforge[3], Aglaé Jézéquel[4,5], Murathan Kurfalı[6,7], Ni Li[8,9], Gabriele Messori[10,7], Joakim Nivre[10,7], Barbara Pernici[11], Niko Speybroeck[3], Stefano Terzi[13], Wim Thiery[8], Bram Valkenborg[8,14], Jingxian Wang[11,12], Shorouq Zahra[10,7], Jakob Zscheischler[1], Jan Sodoge[1,15]

[1] Helmholtz Centre for Environmental Research, Germany
[2] Leipzig University, Germany
[3] UCLouvain Brussels, Belgium
[4] Ecole des Ponts, France
[5] Université PSL, École Polytechnique, Institut Polytechnique de Paris, Sorbonne Université, CNRS, France
[6] RISE Research Institutes of Sweden, Sweden
[7] Swedish Centre for Impacts of Climate Extremes (Climes), Sweden
[8] Vrije Universiteit Brussel, Belgium
[9] TUD Dresden University of Technology, Germany
[10] Uppsala University, Sweden
[11] Politecnico di Milano, Italy
[12] Istituto Universitario di Studi Superiori IUSS Pavia, Italy
[13] Center for Climate Change and Transformation, Eurac Research, Italy
[14] Royal Museum for Central Africa, Belgium
[15] Technopolis Group, Germany

*Corresponding author. E-mail: mariana.brito@ufz.de

**Abstract**

Recent advances in natural language processing (NLP) and large language models (LLMs) have enabled the systematic use of large-scale textual data from news, social media, and reports to create datasets with socio-economic impacts of climate hazards such as floods, droughts, storms, and multi-hazard events. As the field of text-as-data for impact assessment expands, so does its methodological complexity. Yet research remains fragmented, with no clear guidelines for defining what constitutes an impact, handling temporal and spatial biases, and selecting appropriate modeling and post-processing strategies. This lack of coherence limits transparency and comparability across studies. Here, we address this gap by synthesising common practices, describing key challenges specific to the use of text-as-data methods for analyzing socio-economic impact data, and proposing recommendations to address them. By providing guidance on best practices, we aim to support the construction of robust text-derived socio-economic impact datasets that can more accurately inform disaster risk management and attribution studies.

# 1 Introduction

Climate hazards—such as floods, droughts, heat waves, and storms—are perceived as the top-ranked global risk over the next decade according to the World Economic Forum [1]. They disrupt economies [2], compromise human health [3,4], strain food systems [5], and intensify forced migration [6]. To better understand the interplay among climatic and societal factors that drive these losses [7] and, thus, support risk reduction, reliable data on impacts are needed.

Despite the importance of impact information, detailed data are often exclusive to private ownership, such as (re)insurance companies, and when available, are limited to more easily quantifiable consequences at aggregated levels, such as fatalities per event [8] or annual crop losses [9]. As a result, impact data are mostly aggregated and limited to direct, measurable consequences, hampering our understanding of how multiple hazards interact and lead to cascading impacts [10]. While cross-sectoral impact assessments do exist [11,12], they are often based on manual approaches that are time-consuming and difficult to scale or update in near-real-time. Likewise, freely available impact datasets with global, multi-hazard coverage (e.g., EM-DAT [13], DesInventar [14]) are constrained by their reliance on a limited number of sources, focus on large events, and hence lead to missing data and spatio-temporal biases [15,16].

To reduce impact underreporting, scientists are increasingly using natural language processing (NLP) and large language models (LLMs) to extract impact information from unstructured text data [17–20]. These efforts aim to both identify when and where an impact is reported [21] and extract information on its socio-economic consequences [22]. An advantage of these emerging datasets is their potential to increase both temporal and spatial coverage, as the abundance and diversity of text sources such as news articles, social media posts, websites, and reports may, in principle, reduce the likelihood of missing events and enable continuous updating. Moreover, rich-text descriptions can help study indirect impacts that are often overlooked (e.g., mental health and cultural impacts). Thus, this new generation of text-based impact datasets holds the promise for filling critical gaps in impact analysis and monitoring, such as enabling near-real-time updates, improving coverage in underrepresented regions, and supporting the collection of data on multiple hazards.

However, the rapid development of this field has outpaced the establishment of shared methodological standards. As a result, studies differ vastly in what they consider as an “impact” and how its magnitude is computed. For example, some treat any mention of crop disruptions as an agricultural impact [17], while others classify impacts only when quantitative economic losses are provided [22]. As a result, datasets generated by these approaches are not comparable, and cross-study analyses can lead to contrasting conclusions. This complexity creates barriers for scientists entering the field and challenges for reviewers tasked with assessing the dataset robustness. Hence, as with any paradigm shift, there is a need for the research community to establish ground practices to guide the use of NLP and LLMs in impact assessments. Beyond methodological heterogeneity, text-based impact data face acceptance challenges within primarily quantitative assessment communities, where texts are sometimes dismissed as subjective relative to sensor-based measurements [23]. This further underscores the need for reporting and validation standards to build legitimacy across disciplines.

To address this need, this perspective paper makes three main contributions. First, it provides an overview of existing application cases, showcasing how text-based impact datasets can add value beyond existing repositories by addressing gaps such as including indirect consequences, near-real-time monitoring, and sub-national detail. Second, it identifies common challenges across the impact identification and extraction pipeline—from decisions about which impacts to represent to challenges in information extraction, modeling, and post-processing. Third, it provides practical guidelines for scientists, reviewers, and other text-as-data users to improve the quality of text-based impact datasets. The proposed recommendations are grounded in our collective experience spanning geography, engineering, computer science, climate science, epidemiology, and linguistics. Together, they aim to foster more consistent, rigorous, yet adaptable approaches to assessing the societal impacts of climate hazards using text-as-data.

# 2 Text-as-data in socio-economic impact assessments

To ground our effort in providing guidelines for advancing the field of NLP and LLMs for impact assessments, we first reviewed studies that use text-as-data to investigate the socio-economic impacts of floods, droughts, storms, landslides, wildfires, heat waves, or cold spells. This review was not intended to be exhaustive, but to provide a snapshot of common practices in this domain. Based on a Web of Science search followed by manual screening, we identified 64 relevant peer-reviewed studies published between 2016 and 2025 (Fig. 1). Each study was reviewed by at least two co-authors, with disagreements adjudicated through a discussion with a third co-author. The inclusion criteria, search keywords, and review protocol are described in Supplementary Material 1, and the list with all studies is provided in the Supplementary Data.

Results show that social media (n=45), especially Twitter and Weibo, are the most-used data sources, reflecting their relative accessibility prior to the end of the "free API era" [24]. In contrast, the use of news articles (n=13) and institutional reports (n=9) is limited, likely due to copyright constraints [25], high access costs [26], and the lack of standardized repositories of reports.

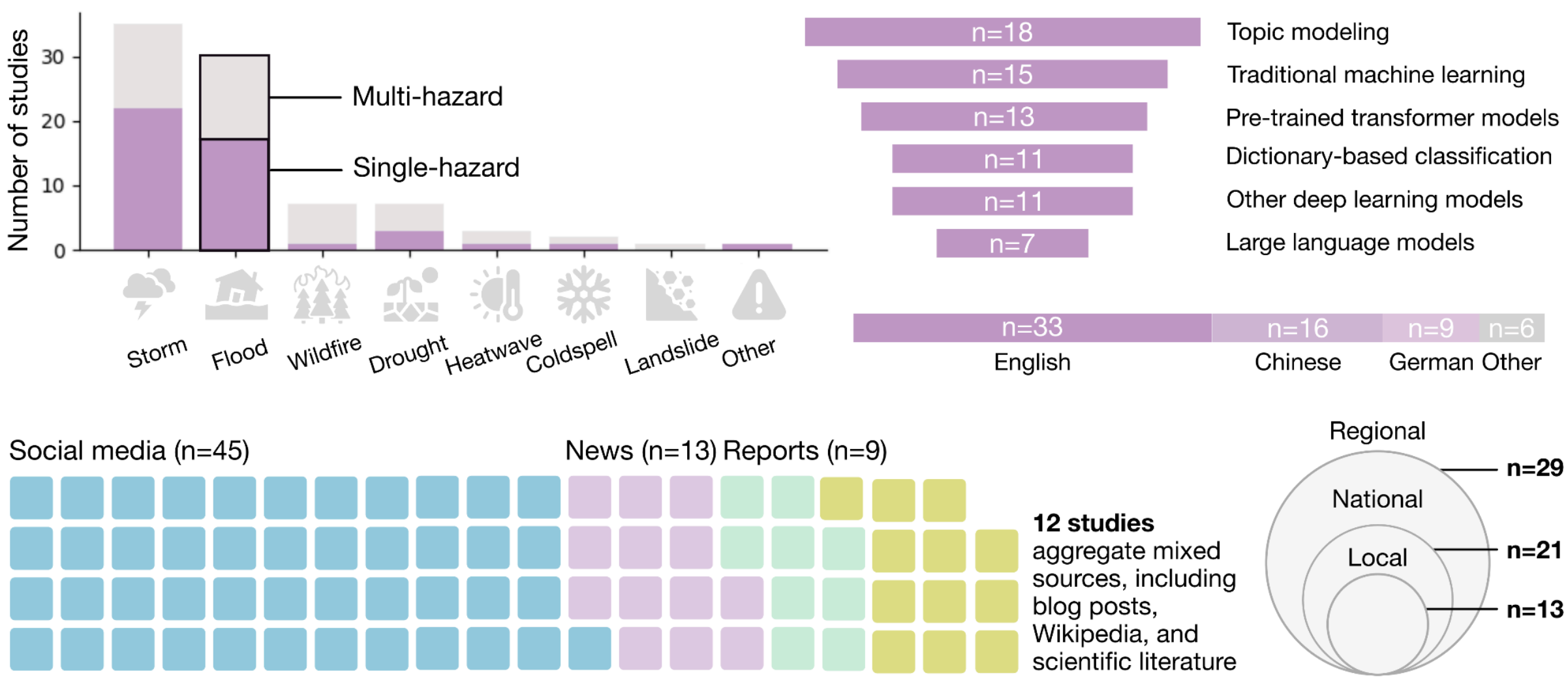

**Fig. 1 | Overview of 64 peer-reviewed studies that use NLP to examine the socio-economic impacts of climate hazards based on text data.**

A broad range of impacts was studied, some of which are often overlooked in traditional impact databases, such as impacts on tourism [27], mental health [28], drinking water [29], and power supply [30]. This illustrates the added value of using text data to monitor impacts: the categories are not constrained by predefined frameworks and can be tailored to the specific research question or to the socio-environmental context of the case study.

About 72% of the studies examined single hazards, with storms (n=35) and floods (n=30) being the most studied types. Conversely, landslides, cold spells, and heat waves received little attention, with fewer than three studies each. Despite the capacity of textual data to capture multiple concurrent or interacting events, multi-hazard analyses remain limited (n = 18).

In terms of spatial scale, NLP was mainly used to study impacts at fine, sub-regional resolutions. Most studies focused on regional-level analyses (n=29), followed by national (n=21) and local (n=13) scales, with no study providing a global assessment. From a temporal perspective, researchers mainly concentrated on a single event (n=38), with a smaller subset (n=24) incorporating multi-year datasets.

Unsupervised NLP methods like topic modeling (e.g., via Latent Dirichlet Allocation[31,32] and BERTopic[33]) were dominant (n=17). In many cases, these studies did not explicitly aim to assess impacts. Instead, topic modeling was used to identify latent themes in the text, some of which corresponded to socio-economic impacts (e.g., [34]). Traditional machine learning methods, including Naïve Bayes [35] and Lasso logistic regression [36], were applied in 16 studies. Despite their relative simplicity, these approaches have demonstrated strong performance in detecting impacts while remaining interpretable. Deep learning-based NLP approaches that capture contextual complexities, such as Transformer-based models (n=11) and other architectures (n=10), were also prevalent. More recently, large language models (LLMs) have also been applied (n=7), for instance, using question-answering approaches to extract information on disasters and their impacts [28,37].

Overall, we identified four main text-based approaches to study impacts (Supplementary Box 1). **Impact presence detection**, used in 13 studies, focuses on identifying whether an impact is mentioned in a text, regardless of its frequency [36]. Building on this, most studies derive **impact salience metrics** (n=42), for example, by counting the frequency of impact mentions [38,39] or by calculating the share of documents describing a given impact type [31,36]. A smaller number of studies focused on extracting **impact narratives** [40], which involve retrieving textual descriptions of reported impacts (n=5). **Impact quantification** was the least used approach (n=4). These methods aim to retrieve numerical values associated with specific impacts, such as the number of fatalities or the amount of economic losses [22].

# 3 Key challenges in text-based climate impact datasets

The promises of using text-as-data for impact assessments are many, including a better understanding of understudied impact types, the provision of data for overlooked regions, and the possibility of continuous data updating. However, our literature review (Fig. 1) reveals that these promises have not yet been completely fulfilled. Turning text into reliable impact datasets requires awareness of its associated challenges. Without careful consideration, using NLP for creating text-based impact datasets can introduce biases and immature practices, leading to unreliable results. To help navigate this emerging field, we outline here key challenges associated with developing text-based impact datasets. Challenges related to NLP research more broadly (e.g., annotation, linguistic aspects) [41–45], misinformation risks [46,47], and gaps in traditional impact datasets [48–51] have been covered elsewhere.

## 3.1 Text corpus selection

The choice of text sources strongly influences the characteristics of text-based impact datasets. With some exceptions (e.g., humanitarian and emergency reports), most text corpora used in impact studies—such as news articles, Wikipedia entries, or social media posts—are not originally designed for impact monitoring. They thus reflect social, cultural, and institutional factors that influence which impacts are reported [52] and whose experiences are documented [53]. Even dedicated impact-monitoring corpora may carry intrinsic biases [54]. The choice of text sources thus inevitably introduces systematic representational, temporal, and spatial biases.

Representational biases arise because different hazards, impact types, populations, and regions receive unequal attention. Social media platforms, for example, overrepresent younger and urban individuals [55]. Thus, disasters affecting rural populations, older people, and individuals with limited internet access may be underrepresented on social media. Scientific literature and news, on the other hand, tend to focus on high-income countries [56]. Coverage also varies by hazard, the speed of onset, and impact salience. For example, droughts often receive less attention than floods and storms [57].

Temporal biases emerge from shifts in data availability and reporting practices. The volume of peer-reviewed literature, digitally available news archives, and social media posts has grown over time [48], leading text-based impact data to show inflated, and potentially even artificial, upward trends due to the increasing volume of text data. Shifts in public awareness (e.g., due to heightened or even reduced concern about climate change) further affect reporting patterns. As a result, increases in reported impacts may reflect changes in reporting behavior rather than an actual increase in the frequency of impacts.

Spatial biases arise from structural inequalities in existing text repositories. The dominance of English-language sources limits the representativeness of impacts in regions such as the Global South [22]. Disasters in Africa, for example, require 45 times more fatalities than those in Europe to receive coverage in Western media [57]. Media outlets also prioritize events relatable to their audiences or aligned with shared economic interests or cultural proximity [58,59]. Even within countries, coverage is uneven. Urban areas, with higher media presence, are often better represented than rural or remote regions [60]. This creates a critical "data gap" in which some of the most exposed areas remain poorly documented.

## 3.2 Event and impact definition

The interpretation of text-based impact datasets depends critically on how "events" and "impacts" are defined. As training and evaluating NLP models often rely on human-annotated data, ambiguous or arbitrary definitions introduce uncertainty that propagates into model evaluation and downstream analyses. Such ambiguity also makes it difficult to compare results with existing datasets (e.g., EM-DAT [13]) or across studies analyzing the same event.

Different needs for event definitions exist depending on the study goals. Studies examining the salience of impacts in discourse often consider a continuous timescale independent of any particular event (e.g., [17,36]). As such, they aggregate impacts over fixed time windows (e.g., weekly) across districts, states, or countries. By contrast, studies that quantify impacts require a clear definition of the event's boundaries (e.g., [22]). This includes specifying the event's start and end dates, and the area it affects. Different choices can influence impact estimates: including a broader time window may capture indirect and delayed effects, whereas a narrower definition may focus only on immediate damages. Without clear spatial and temporal boundaries, impact estimates become difficult to interpret or compare.

Defining impact categories poses similar challenges. Decisions about what constitutes an impact and which impact types to include shape the dataset's usefulness for other projects. Textual descriptions of impacts vary widely. They can range from explicit statements to subtle or indirect references. For example, a news article may state explicitly, "a decrease in mental health was observed following the hurricane", while another may indirectly describe a similar impact, saying that "people became anxious due to the hurricane". As such, boundaries between relevant and irrelevant content are often blurred by individual interpretations. This complicates both data annotation and impact assessment using NLP models. However, it should be noted that there are legitimate sources of disagreement among human annotators on what constitutes an impact, especially when the text has lost its structure (e.g., by being split into paragraphs or sentences) and the textual descriptions of impacts are vague.

Choices about how impact classes (e.g., health, infrastructure) are defined further influence the dataset's comparability and the impacts that appear dominant. For example, some impact classes are conceptually highly similar (e.g., "displacement" and "homelessness") and therefore difficult to distinguish using automated text processing. The lack of clear boundaries can lead to duplicate counts or classification errors.

## 3.3 Impact information identification and extraction

Societal impacts reported in texts are often described in an ad hoc manner, embedded in rich narrative contexts, which complicates both impact annotation and their extraction using NLP. This includes, for instance, information regarding temporal dimensions (i.e., past historical disasters, current events, or future impacts [61]), as well as details on hypothetical impacts (i.e., economic losses are expected). As a result, these difficulties may be propagated to NLP models trained and evaluated on such data.

Extracting structured temporal data on the impacts also presents challenges, as texts often rely on relative references (e.g., "recently" or "for several weeks"). These must be correctly anchored to the text publication date to determine the impact timing [19]. Without adequately handling these nuances, text-based impact databases may misrepresent the severity and timing of reported impacts.

Particular issues arise depending on the study goals. For example, numeric impact data, such as economic losses or the number of affected people, are rarely reported in standardized formats in texts. Even common impact metrics, such as crop yield losses, are reported using different metrics, including yield volume, percentage decrease in harvest, or monetary value. Texts can also contain conflicting numbers. For example, reported victim counts often differ from estimates that account for excess mortality or climate-change attributable deaths [62]. Therefore, extracting harmonized figures or comparing them to other datasets is difficult [20].

The choice of NLP methods and data types further shapes these limitations. Supervised classification models struggle to accurately classify rare impacts with little training data or when impacts are ambiguous even for human annotators (e.g., societal conflicts). Simpler approaches based on keywords or bag-of-words fail to capture nuanced temporal and contextual information from texts. Unsupervised approaches such as topic models present a different challenge: it is hard to align topics with predefined impact typologies, even when using seeded models [63]. These limitations are amplified by differences in text sources: while news articles are professionally edited, social media posts use more colloquial language and may contain misinformation [46], requiring additional steps to remove unreliable content.

## 3.4 Impact geolocation

Mapping impacts to their corresponding locations is crucial for spatial analyses. The process typically involves three steps: toponym recognition (detecting location names), toponym disambiguation (resolving toponyms to geographic units or coordinates using a reference database), and identifying the actual location where the impact occurred among the extracted toponyms. While toponym identification and disambiguation methods are well-established [64,65], they are not flawless [43,66–69]. Even the current best-performing tools fail: they may miss valid locations, fail to match detected toponyms to coordinate databases, incorrectly resolve ambiguous toponyms, and select the wrong event location among all extracted toponyms.

Additional biases arise from the geographic coverage of available off-the-shelf tools, which tend to overrepresent places in wealthier and English-speaking regions [70]. This uneven coverage introduces spatial biases and inaccuracies [71], thereby affecting conclusions obtained. For example, texts discussing disasters often refer to natural features (e.g., rivers, mountains), infrastructure (e.g., roads, dams), and points of interest (e.g., parks); however, these features are often underrepresented or inconsistently labeled in geocoded databases and geoparsers (e.g., [72–74]), which often prioritize administrative units, such as countries, states, and cities [43].

Ambiguity increases when multiple impacts and relevant locations are mentioned in the same text. For example, a text may state that "the flood destroyed a bridge in Bonn, and left 100 children without school in Cologne". Simply extracting all detected place names risks assigning infrastructure damage and school disruption to both cities, thereby inflating impact estimates. The issue intensifies for longer texts, which may reference several locations unrelated to the impact (e.g., past disasters elsewhere). Researchers thus face a trade-off: considering the entire paragraph or document risks detecting irrelevant locations, whereas restricting the analysis to sentences around the detected impacts may exclude relevant contextual information needed to map the impact correctly. LLMs are often proposed as solutions to these cases, but these models can also struggle with spatial attribution when multiple locations are reported [75].

## 3.5 Data post-processing and dataset updating

Once socio-economic impacts and their locations are extracted from a text, creating a reliable and updatable impact dataset requires a series of post-processing and validation steps. This includes standardizing data, correcting biases, and removing duplicate entries. Such steps are critical because automated impact-extraction processes introduce errors that cannot be manually verified at scale.

When using salience metrics, post-processing includes defining how impact mentions are aggregated into a metric of impact severity. Such approaches assume that more frequently reported impacts are more relevant [76]. Yet reporting frequency may reflect media attention, editorial priorities, or spatial [77] and temporal [78] biases rather than impact severity. Thus, post-processing needs to address distortions causing reporting volume and impact severity to misalign. However, correcting for these biases is difficult in the absence of ground-truth observations, which is often the case for impact data.

For research focused on quantifying impacts, once an impact data point has been identified and extracted from the text (potentially verbatim from the source), it must be standardized to enable comparison. A challenge lies in handling ambiguous numerical expressions, such as “hundreds of deaths”. Converting such statements into single values (e.g., 500) or ranges (e.g., 100–999) introduces biases [22]. Similarly, standardizing monetary values and adjusting for inflation involves non-trivial choices [79]. A further challenge lies in distinguishing between duplicate estimates, genuinely new impacts, and updated figures. For example, while humanitarian reports are typically updated only a few times per event, news and social media provide new information continuously, often with changing values (e.g., fatalities counts), making it difficult to identify a “final” number.

Finally, updating near-real-time impact datasets is operationally challenging because text sources are widespread and scrapers often break due to changes in website structure, access restrictions, or anti-scraping measures. Therefore, the data collection pipelines need to be regularly adapted [80]. Moreover, proprietary datasets, such as curated news aggregations or access to social media APIs, are often expensive and inaccessible to many researchers. Data sharing for certain text types is further limited by copyright restrictions, raising ethical and legal considerations.

# 4 Recommendations for creating text-based impact datasets

In light of the challenges discussed, we, as a community of researchers working at the intersection of climate science, disaster risk, and NLP, have collaboratively developed recommendations to support developing text-based impact datasets (Fig. 2). Supplementary Material 2 provides a detailed checklist. These recommendations are intended as a reference for students, researchers, and practitioners seeking to improve the transparency and credibility of text-based impact datasets. While primarily directed at dataset creators, they can also help end users and reviewers assess the quality of the impact dataset. We stress that they should serve as a reference rather than rigid requirements, since the appropriate design choices depend on the dataset's purpose, use case, and available text sources.

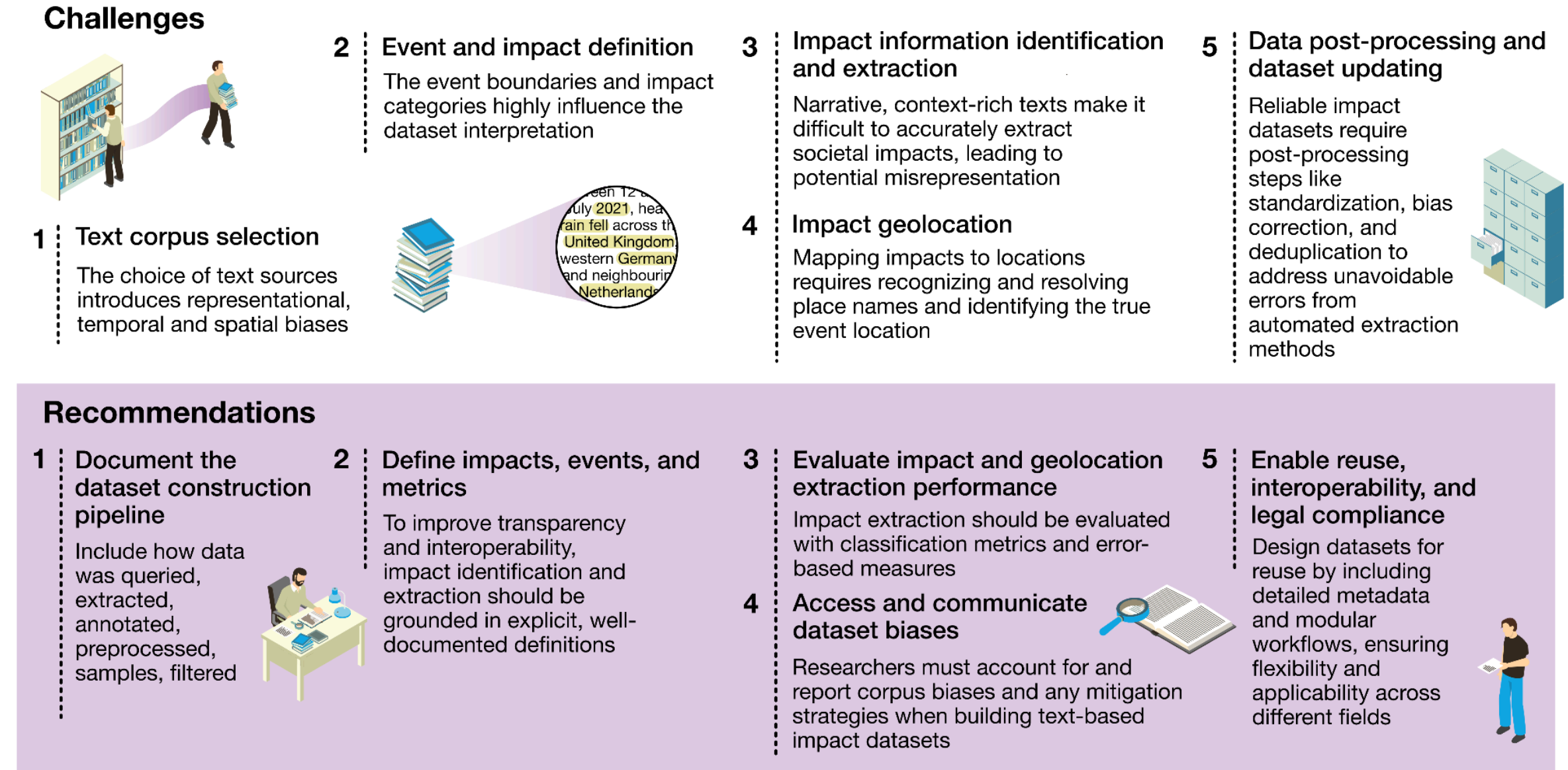

**Fig. 2 | Recommendations for creating reliable text-based impact datasets.**

## 4.1 Document the dataset construction pipeline

As a first step, we recommend adhering to responsible practices for dataset development [44] and the FAIR principles (i.e., findability, accessibility, interoperability, and reusability) [81,82]. All stages of dataset construction should be clearly documented. This includes how text data was queried, extracted, annotated, preprocessed, sampled, filtered, or otherwise transformed. The documentation should describe the steps taken, their order, and the rationale, as well as all included or computed metadata. This is critical to ensure that impact extraction workflows are reproducible and easily evaluated by others.

Researchers should report aspects such as the guiding theory, search and selection strategies, inclusion and exclusion criteria, coverage (temporal, spatial, linguistic, genre, etc.), known and potential biases, and any other information that may be useful to others. These aspects should be disclosed, as users from other fields or outside of academia may not have the necessary background to identify the consequences of each choice. Existing documentation frameworks, such as the *datasheet for datasets* by Gebru et al. [83] and related *data statement* guidelines for text-as-data (e.g., [84,85]), can provide useful templates for this purpose.

To facilitate reuse, open-source code should accompany the impact dataset, so that it can be reproduced, inspected, and improved by other users (e.g., to correct flaws or to extend the dataset). Depending on copyright regulations, raw text data should be included. Provenance information (e.g., exact source and retrieval timestamp) must be logged for every included document and extracted impact instance to improve traceability [86], reanalysis, and facilitate the adaptation of the results to other applications. Moreover, unique identifiers should be assigned to each observation following best practices (e.g., [87]) to facilitate referencing.

**At minimum, studies should report data sources, retrieval strategy, time period, languages covered, preprocessing and filtering steps, annotation procedures, and provenance information.**

## 4.2 Define impacts, events, and metrics

To improve transparency and interoperability, impact identification and extraction should be grounded in explicit, well-documented definitions. The metadata accompanying impact metrics should specify the temporal and spatial scope of the data and whether the impact is linked to single or multi-hazard events, thereby guiding the aggregation and comparison across multiple entries [88].

Full compliance with a single standard may not always be feasible (or desirable), owing to mismatches regarding the dataset's purpose, user needs, or the corpus’s information content. Nonetheless, existing disaster loss reporting standards and terminology, including international (e.g., UN-related frameworks [89]), regional [90,91], or national frameworks [92] and established disaster loss databases, such as EM-DAT or NatCatServices [13,49,50], can serve as valuable reference points. For impact salience measures, researchers should always report raw counts alongside normalized values, allowing end users to test different assumptions.

As impact data is used across multiple scopes and disciplines, the definitions should ideally be co-developed with domain experts and end users. These definitions must be robust enough to handle edge cases. This includes, for instance, criteria for distinguishing between past and hypothetical impacts and determining which losses (e.g., property damage, environmental degradation) are considered.

**At minimum, studies should report how events and impacts are defined, the temporal and spatial unit of analysis, whether impacts are direct or indirect, and the exact metric used.**

## 4.3 Evaluate impact and geolocation extraction performance

Evaluating the extraction of impact data requires addressing its multidimensionality. For binary classification tasks, standard metrics such as precision, recall, and F1 score, evaluated on manually annotated data, can be used to assess whether impacts are correctly identified. For studies that extract numeric values (e.g., fatality counts), researchers need to evaluate both the correctness of the impact type and its associated magnitude, using metrics such as exact match, relative error, or tolerance-based thresholds.

For geolocation tasks, ideally, more than one tool should be tested and their results compared as even off-the-shelf tools have flaws and require quality control. The practical guide by [67] presents an overview of evaluation best practices in geoparsing. Also, since good performance on other datasets may not generalize to the selected impact documents, annotating a sample of one's own data to serve as a gold standard is needed to estimate performance and conduct error analysis. Annotators ought to follow a proper coding scheme to assess geolocation performance. A prior decision on the appropriate administrative levels is needed to ensure that all instances are mapped to equivalent geographic units.

Multiple evaluation metrics should be computed as each metric is a proxy for a different construct of interest. While the nature of each geoparsing subtask calls for its own metrics, there are some conventions [66–68]: toponym recognition relies mainly on precision, recall, and F1 score, whereas toponym resolution typically assesses area under the curve, location overlap, and mean error distance. Computing the proportion of exact matches between geolocated region names can be appropriate when the target is a large administrative unit (e.g., a country). However, finer-grained geolocation (e.g., a city) requires additional measures, such as assessing the degree of overlap between the resolved and true locations. In any case, the exact metric variations (e.g., binary, macro-, or micro-F1 score) should be reported.

To handle ambiguous cases in which multiple impacts and locations are extracted, researchers can use spatial clustering or identify spatial centroids [93,94]. While clustering can help filter out distant geographic outliers, it cannot fully resolve ambiguity when locations are close together. Therefore, additional strategies are needed, such as weighting locations by contextual relevance, training domain-specific location-extraction algorithms, integrating geographical dictionaries with machine-learning models, or applying rules based on syntactic proximity. Ultimately, combining automated methods with careful human oversight using a human-in-the-loop approach can improve confidence in this task.

**At minimum, studies should report the target spatial unit, the metrics used to assess impact information and geolocation extraction performance, and the geoparsing tools used.**

## 4.4 Access and communicate dataset biases

When building text-based impact datasets, researchers need to understand the potential biases each text corpus carries, as they shape which impacts, hazards, and perspectives are captured. No corpus is neutral: selection into visibility (what gets written), selection into retrieval (what we capture), and selection into labeling (what we annotate) jointly shape which impacts are reported. Researchers should make these assumptions explicit and auditable to avoid mistaking changes in reporting for changes in impacts. Hence, before collecting text from any source, its quality, biases, and potential for misinformation should be verified. A systematic approach to evaluating data quality is provided by [95].

To reduce corpus-related biases, we recommend using multilingual and diverse sources, such as multiple news outlets, Wikipedia, and peer-reviewed articles. Also, underrepresented sources (e.g., grey literature, government reports) should be included when possible to offset Global North dominance. Using multiple heterogeneous sources can also support information triangulation, strengthening confidence in text-derived data.

Since annotation procedures also influence dataset reliability and biases, relying on multiple annotators should be done when possible. Agreement should not be "manufactured" just to achieve a good inter-annotator metric, as it often reflects genuine ambiguities [96]. Interdisciplinary discussions could help develop meaningful guidelines to address fuzziness in data annotation.

Once the impact dataset is established, scholars may employ post-processing strategies depending on the use case. This includes, for instance, removing near-duplicates (i.e., documents reporting on the same impact the same day) [36] or normalizing the impact counts by the total population exposed [97]. Furthermore, to correct for temporal biases when using impact salience metrics, a common strategy is to normalize impact counts by the total number of published documents in a given time period [36] or to consider impact shares over time rather than counts [33].

Persistent biases should be evaluated through external comparison with other impact datasets. Results can be benchmarked against established disaster databases as well as national or local impact statistics. However, comparing event counts may simply reveal differences in definition and resolution between datasets. A more thorough completeness evaluation must therefore account for event overlaps. Transparent discussion of discrepancies in impact and occurrence is essential. It should be noted that datasets such as EM-DAT should serve as calibration points rather than the "ground truth" as it is very hard to define what the "ground truth" is, even for impacts from well-observed events [98]. Finally, engaging panels of end users may further support the identification of systematic biases and provide guidance on complex classification issues.

**At minimum, studies should report likely representational, temporal, spatial, and linguistic biases, as well as any mitigation or post-processing strategies applied.**

### 4.5 Enable reuse, interoperability, and legal compliance

Impact datasets should be constructed with generalizability and a long-term view in mind: a collection too narrowly tailored to a single study may not be reusable for other purposes. To make the data useful to others, it needs to include all available metadata so that users can decide for themselves what they need. Furthermore, text processing workflows should be modularized so that users can skip or add processing steps as needed.

Legal considerations are also central to ensuring dataset usability. Copyright and licenses should be considered when collecting, and releasing text-based impact data (see [99] for an overview). Researchers must ensure they have the right to collect data and confirm that their intended use is permitted by the data source provider. Scraping webpages, for example, is not always legal and can infringe copyrights. Furthermore, every released dataset must be accompanied by a license that specifies permitted uses. A common option is the Creative Commons licenses [100], which offer varying levels of permissibility. Juridical assistance may be needed to ensure that the redistribution of derivatives from various data sources is unified into a single release under a suitable license. Within these constraints, we recommend making databases open access whenever possible, at minimum for academic research and other non-profit applications.

Finally, collaboration with scientists from different fields can help improve the database construction process through interdisciplinary discussions. Such efforts can also support the identification of gaps that text-based databases could fill. For example, collaboration with climate scientists may support the integration of hazard information (e.g., flood extent maps, precipitation anomalies, drought indices) with text-based impact datasets. Linking these sources can help guide document retrieval toward periods of known extreme climate events and support tasks such as geoparsing and temporal attribution. However, such integration requires caution, as the spatial and temporal footprints of hazards and their impacts are not the same.

**At minimum, studies should report licensing conditions, available metadata, versioning practices, and any restrictions on data sharing or reuse.**

## 5 Summary and future perspectives

Using text-as-data offers a powerful way to track climate impacts that are hard to capture with traditional data types, especially indirect, local, or long-term consequences. NLP-based approaches can enable analysis at granular geographic scales and in near-real-time. However, our literature review (Fig. 1) shows that although progress has been made in examining often-overlooked impact types, the promise of NLP for building longitudinal, updatable, multi-hazard databases remains unrealized. A problem is that the very qualities that make texts so useful (i.e., their open-ended nature, multilingualism, and rich contextual detail) also introduce biases. Therefore, text should be treated as a combination of evidence of impacts and personal impressions of what people noticed, said, or prioritized.

To make progress in the field, researchers need to agree on useful standards. Shared definitions can make it easier to compare across studies. Setting minimum reporting and documentation guidelines may make it easier to review and replicate results. Datasets should also record where the impact information came from, how certain it is, what biases exist, and how this information has changed over time, which helps avoid mistakes such as double-counting. At the same time, incorporating human-in-the-loop approaches, where expert judgment is used to validate, refine, and interpret model outputs, can help ensure contextual accuracy.

Still, critical challenges remain. Despite the current hype, NLP, LLMs, and AI methods are not faultless, and their good performance must not be taken for granted. Natural language is complex, featuring polysemy, figures of speech, irony, negation and related phenomena. These characteristics challenge automated parsing and even human interpretation. Language is more than mere "data": our idealizations about how it manifests in texts are not always correct, and its mechanisms are not summarizable into simple processing rules. Additionally, some tools used to extract information from text are not deterministic, hindering reproducibility. Many of such tools are not open-source – notably several state-of-the-art LLMs – which limits reproducibility even when a clear, documented pipeline is provided. For example, older model versions on which a dataset is based may no longer exist.

Looking ahead, near-term progress will likely come from a combination of methodological experimentation and collaborative practices. Researchers can systematically compare results across different text sources to identify impacts where most sources agree, as well as reporting gaps. Establishing shared benchmark impact datasets for multilingual evaluation tasks would also allow the community to assess impact extraction and geolocation methods under comparable conditions. Finally, integrating external reference datasets, such as disaster loss databases, population exposure statistics, or hazard information, is essential for contextualizing textual signals, calibrating extracted impact estimates, and detecting inconsistencies. These steps move beyond basic reporting standards toward a more robust research ecosystem, where text-derived impact datasets are not only transparent but also comprehensive, comparable, extensible, and analytically reliable.

**Acknowledgements**

**Funding**

**ST** was partially funded by the project PE00000005_RETURN: "Multi-Risk sciEnce for resilient communities undeR a changing Climate" PNRR, under the Missione 4, Componente 2, Investimento 1.3, call n. 341 del 15/03/2022 CUP: D53C22002510002 declaration. **BP** was partially funded by the PNRR-PE-AI "FAIR" project funded by the NextGenerationEU program. **GM** was partially funded by the projects Swedish Research Council Vetenskapsrådet grants nr. 2022-06599 and 2022-03448. **BV** was partially funded by the project GuiDANCE Project, Belgian Science Policy Office, FED-tWIN Programme n° Prf-2019-066-GuiDANCE. **WT** acknowledges funding from the European Research Council (ERC) under the European Union's Horizon Framework research and innovation programme (grant agreement No 101124572; ERC Consolidator Grant 'LACRIMA').

**Conflict of interest:** The authors declare no competing interests.

**Data availability:** The systematic review results are available in the Supplementary Data

**Code availability:** No custom code was developed or used for the analyses conducted in this study.

**Author contribution**

Conceptualization: MdB (lead), JS (contributor)

Methodology: MdB

Validation: MdB (lead), TMNC, JS, NL, JW, SZ, BV, MK

Investigation: MdB, TMNC, JS, NL, JW, SZ, BV, MK

Data curation: MdB

Writing – original draft: MdB (lead), JS (co-lead), BM (co-lead) and all co-authors were contributors

Writing – review & editing: MdB (lead) and all co-authors were contributors

Visualization: TC

Project administration: MdB

**Supplementary Material 1: Literature review methodology**

To identify studies that use text-as-data to assess the socio-economic impacts of climate-related hazards, we conducted a systematic literature review. We searched the Web of Science (WoS) database on 04 July 2025 using a query (Table S1) combining three groups of keywords: (i) climate hazard, (ii) natural language processing and text-as-data methods, and (iii) socio-economic impact terms. The query covered the full available WoS publication record up to June 2025, without imposing a minimum start date.

**Table S1 | Search keywords used to identify peer-reviewed articles investigating disaster impacts using NLP and text-as-data methods**

| Concept | Query |
|---|---|
| Hazard | (multi-hazard OR "several hazards" OR "compound hazard*" OR drought* OR "dry spell*" OR dryness OR "water shortage*" OR "water insecurit*" OR "scarse water" OR "groundwater depletion*" OR flood* OR inundation* OR "glacial lake outburst" OR storm* OR superstorm* OR windstorm* OR snowstorm* OR blizzard* OR derecho OR "winter storm*" OR hail OR cyclone* OR thunderstorm* OR tornado* OR "storm surge*" OR "storm tide*" OR hurricane* OR "hail storm*" OR typhoon* OR "strong wind*" OR "wind gust*" OR "extreme rain" OR "extreme precipitation" OR "heavy rain" OR "heavy precipitation" OR cloudburst* OR heat!wave* OR heat episode* OR "hot weather" OR "high temperature*" OR heatspell* OR hotspell* OR "heat stress" OR "extreme temperature*" OR "extreme heat" OR cold!wave* OR "cold snap" OR "arctic snap" OR "severe winter condition*" OR "cold spell" OR "low temperature*" OR "cold weather" OR "extreme cold" OR landslide* OR "rock fall*" OR mudslide* OR "mass movement*" OR "forest fire*" OR wild!fire* OR land!fire* OR bush!fire* OR "wildland fire*") |
| NLP methods | "natural language processing" OR NLP OR "text-as-data" OR "textual analysis" OR "text mining" OR "text embeddings" OR "large language models" OR "LLMs" OR "named entity recognition" OR NER OR "semantic analysis" OR "text classification" |
| Impact terms | (impact* OR consequence* OR effect* OR damage* OR loss* OR violen* OR crime* OR war* OR conflict* OR dispute* OR unemploy* OR poverty OR income OR “water scarcity” OR “water supply” OR “water availability” OR “lack of water” OR “hydrological stress” OR “drinking water” OR (water AND (chlorophyll OR nitrogen OR phosphorus OR quality OR pollution OR heavy metal* OR pesticide*)) OR algae?bloom OR (food AND (security OR supply OR food production)) OR famine OR livestock OR cattle* OR (animal AND (well-being OR husbandry OR welfare OR nutrition)) OR fishery* OR aquaculture OR fish stock OR health* OR well?being OR ill OR illness OR disease* OR syndrome* OR infect* OR medical* OR disabilit* OR death* OR fatalit* OR died OR casualties OR “loss of life” OR injur* OR infectious disease* OR cholera OR giardiasis OR cryptosporidiosis OR leptospirosis OR "(obes* OR over?weight OR under?weight OR hunger OR stunting OR wasting OR undernourish* OR undernutrition OR anthropometr* OR malnutrition OR malnour* OR anemia OR anaemia OR ""micro?nutrient*"" OR diabet*)" OR mental OR depress* OR *stress* OR anxi* OR ptsd OR psycho* OR psychiatric* OR *trauma* OR post-traumatic OR suicide* OR solastalgi* OR “air quality” OR “air pollution” OR PM2.5 OR “fine particulate” OR asthma OR displacem* OR relocation* OR migration OR refugee* OR homeless* OR emergency shelter OR bridge* OR road* OR highway* OR train* OR transport* OR rail* OR ship OR mobility OR ((water OR waste?water) AND treatment plant*) OR sewage* OR sewer* OR sewerage* OR waste OR landfill OR hospital* OR care clinic* OR emergenc* OR pharmac* OR digital infrastructure OR communication infrastructure OR ((mobile OR *phone OR internet) AND (network* OR system*)) OR energy OR electricity OR heating OR gas supply OR biogas OR ((wind OR hydro OR nuclear OR coal OR thermal) AND power) OR propert* OR house* OR building* OR infrastructure* OR (macroeconomic AND loss) OR economic assets OR capital OR companies OR business* OR industr* OR commerce OR crop* losses OR crop yield* OR crop quality OR crop failure OR yield loss* OR agriculture OR forest dieback OR forest damage OR tree vitality OR tree growth OR tree dieback OR forestry OR die?off OR tourism OR tourist* OR hotel* OR museum* OR culture OR cultural OR recreation*) |

The query returned 458 peer-reviewed articles. Studies were considered relevant if they met these inclusion criteria: (1) they addressed one or more climate-related hazards, including floods, droughts, storms, landslides, wildfires, heatwaves, or cold spells; (2) they used text as a primary data source (e.g., news articles, social media, or official reports); (3) they assessed specific socio-economic impacts such as mortality, economic losses, displacement, or infrastructure disruption; and (4) they applied computational text analysis methods, including natural language processing and large language models.

We excluded (1) review articles, opinion pieces, conference proceedings, or otherwise not peer-reviewed documents; (2) articles not published in English; (3) articles that focused solely on hazard occurrence without assessing specific impacts; (4) articles that examined public sentiment or perception without extracting concrete impact information; or (5) articles that relied primarily on manual qualitative coding without employing computational methods.

An initial screening was conducted by the first author based on titles and abstracts to identify potentially relevant studies. A total of 136 articles deemed relevant at this stage were then distributed among the co-authors MdB, TMNC, JS, NL, JW, SZ, BV, MK for full-text assessment. Each article was reviewed by two authors. In cases of disagreement (n=14), discrepancies were resolved through consultation with an additional co-author. Of the 136 articles reviewed in detail, 64 met all inclusion criteria and were included in the final analysis (Figure S1).

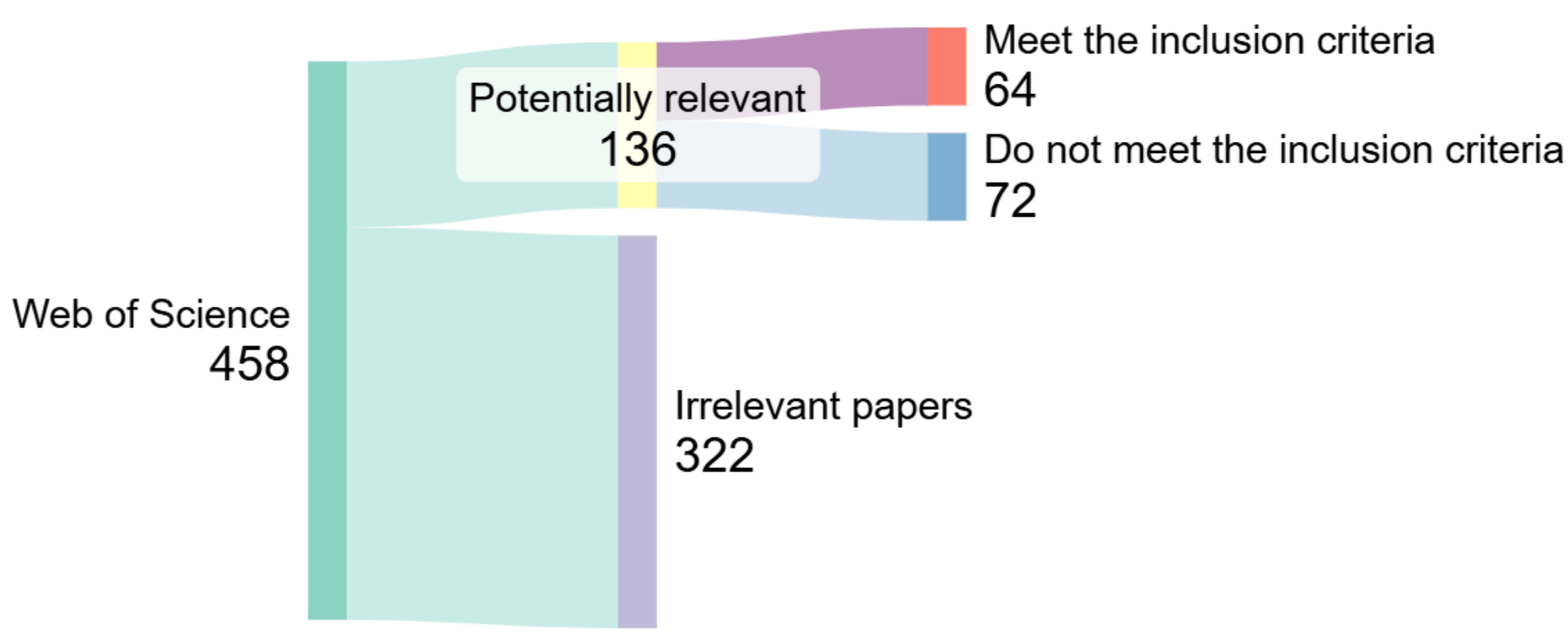

**Figure S1 | Sankey plot with the systematic review search strategy.**

**Supplementary Material 2: Recommendation checklist for constructing text-based climate impact datasets**

## Documenting the dataset construction

**At minimum, studies should report data sources, retrieval strategy, time period, languages covered, preprocessing and filtering steps, annotation procedures, and provenance information.**

☐ 1. The authors are familiar with responsible practices for dataset development and the FAIR principles.

☐ 2. All stages of the dataset construction have been documented:

☐ 2.1. how text data sources were selected;

☐ 2.2. how text data was queried;

☐ 2.3. how text data was extracted;

☐ 2.4. how text data was preprocessed or otherwise transformed;

☐ 2.5. how text data was filtered;

☐ 2.6. how text data was sampled;

☐ 2.7. how text data was annotated;

☐ 3. Relevant methodological details and dataset characteristics have been documented:

☐ 3.1. the guiding methodology and theory;

☐ 3.2. the inclusion and exclusion criteria;

☐ 3.3. the coverage (e.g., temporal, spatial, linguistic, genre);

☐ 3.4. the structure and description of available and extracted metadata;

☐ 3.5. known and potential biases and gaps;

☐ 3.6. any additional information that may be useful to others.

☐ 4. Data management and reproducibility measures have been implemented:

☐ 4.1. the dataset is maintained under version control;

☐ 4.2. if legally permitted, raw data have been included;

☐ 4.3. links to raw sources have been retained when information or fragments were extracted;

☐ 4.4. provenance information (e.g., source, timestamp) has been logged to enable traceability;

☐ 4.5. each record has been assigned a unique identifier to facilitate referencing;

☐ 4.6. the dataset has been accompanied by modularized and reproducible code.

## Providing definitions and metrics for socio-economic impacts

**At minimum, studies should report how events and impacts are defined, whether impacts are direct or indirect, the temporal and spatial unit of analysis, and the exact metric used.**

☐ 1. Robust impact definitions have been developed:

☐ 1.1. relevant conceptual frameworks have been used as reference points;

☐ 1.2. impact definitions have been co-developed with domain experts and end users;

☐ 1.3. impact definitions have been designed to be sufficiently robust to handle edge cases (e.g., past and hypothetical impacts, inclusion of specific losses);

☐ 1.4. decisions on how to distinguish between similar impacts have been documented.

☐ 2. Impact metrics have been accompanied by metadata specifying their temporal and spatial scope and whether impacts were linked to single- or multi-hazard events;

☐ 3. Impact metrics have been accompanied by raw counts, with and without normalization.

## Evaluating the attributed impact locations

**At minimum, studies should report the geoparsing tools, the target spatial unit, the evaluation dataset, and the metrics used to assess performance.**

☐ 1. Clear procedures have been conducted to resolve spatial ambiguity and assess the reliability of the geolocated impacts:

☐ 1.1. ambiguous cases (e.g., multiple impact locations) have been handled;

☐ 1.2. all instances have been mapped to equivalent geographic representations;

☐ 1.3. likelihood, imprecision and uncertainty measures are provided;

☐ 1.4. every design decision on how measures of uncertainty or location overlap have been considered, and whether exact or partial matches were adopted, have been clearly disclosed

☐ 2. Human oversight and annotation procedures have been implemented:

☐ 2.1. automated methods have been combined with careful human oversight to improve confidence;

☐ 2.1. a sample of one's own data has been annotated to serve as a gold standard for performance estimation and error analysis;

☐ 2.3. annotators have followed a proper coding scheme considering key properties of the geolocation task performance.

☐ 3. Evaluation procedures have been conducted and reported to ensure that the data is reliable for its purpose:

☐ 3.1. multiple evaluation metrics have been computed to provide a comprehensive overview of performance;

☐ 3.2. both conventional and, if applicable, custom evaluation metrics have been reported and their computation and variations have been clearly explained;

☐ 3.3. if possible, more than one tool has been tested and compared.

## Investigating, documenting, and reducing biases

**At minimum, studies should report likely representational, temporal, spatial, and linguistic biases, as well as any mitigation or post-processing strategies applied.**

☐ 1. The authors have devoted efforts to understand and report the potential biases each text corpus carries, including:

☐ 1.1. visibility (what was represented);

☐ 1.2. selection and retrieval (what was captured);

☐ 1.3. labeling (what was annotated).

☐ 2. Source credibility and data quality have been assessed and ensured:

☐ 2.1. the quality, biases, and potential for misinformation in the original data sources have been verified;

☐ 2.2. as far as possible, multilingual, diverse, heterogeneous, and underrepresented sources have been included;

☐ 2.3. data quality has been assessed according to relevant dimensions (accuracy, representativity, completeness, and relevancy).

☐ 3. Annotation quality and consistency have been ensured:

☐ 3.1. annotation has been performed by more than one annotator;

☐ 3.2. annotation agreement has not been artificially induced; instead, genuine disagreements have been inspected and discussed;

☐ 3.3. interdisciplinary discussions have helped develop meaningful guidelines to address ambiguity in data annotation.

☐ 4. Bias identification and mitigation strategies have been implemented and documented:

☐ 4.1. post-processing strategies (e.g. removing near-duplicates and normalizing impact metrics) have been carefully designed and documented;

☐ 4.2. temporal biases have been normalized, if applicable;

☐ 4.3. persistent biases have been evaluated through external comparison with other impact datasets;

☐ 4.4. all assumptions have been made explicit and auditable;

☐ 4.5. discrepancies have been transparently discussed;

☐ 4.6. panels of end users have been engaged to support the identification of systematic biases and to guide complex classification issues;

☐ 4.7. results have been benchmarked against established disaster databases, as a means of calibration and not a ground truth.

## Making results reusable and interpretable for multiple communities

**At minimum, studies should report licensing conditions, available metadata, versioning practices, and any restrictions on data sharing or reuse.**

☐ 1. Long-term, interdisciplinary, and flexible usability have been considered:

☐ 1.1. the dataset has been constructed with generalizability and a long-term view in mind.

☐ 1.2. the dataset construction process has been informed by collaboration with scientists from different fields through interdisciplinary discussions.

☐ 1.3. the dataset has included varied and comprehensive metadata to meet the needs of different fields.

☐ 1.4. text processing workflows have been modularized so that users can skip or add processing steps as needed;

☐ 2. Copyright and licenses have been considered when collecting, releasing, and redistributing the datasets:

☐ 2.1. scrapping has been performed without any copyright infringement;

☐ 2.2. the dataset has been accompanied by a license that specifies permitted uses, e.g.

## Supplementary Box 1: Approaches used for studying climate impacts based on text-as-data methods

The landscape of text-based impact datasets is diverse. Based on a close reading of the 64 peer-reviewed studies, we identified four key approaches. These differ primarily in the type of information the studies seek to capture, ranging from explicit quantitative metrics to qualitative descriptions of impacts.

**Impact presence detection:** This approach determines whether a specific impact is mentioned in a text, regardless of how often, in a binary fashion. For example, a classification model might predict whether an article mentions crop failure. This approach is useful for preliminary analyses or when the text dataset is too small to compute salience metrics.

**Impact salience metrics:** This approach identifies mentions of impacts and quantifies their prominence in a given text or corpus. Salience can be measured using raw counts (e.g., the frequency with which sentences mentioning "houses were damaged" appear) or normalized measures (e.g., the proportion of documents that mention a specific impact in the sample).

**Impact narratives:** This category encompasses studies that collect textual descriptions of impacts. Narrative methods can highlight how impacts are framed and contextualized. These qualitative insights are valuable for understanding the nuances and details of impacts.

**Impact quantification:** This approach focuses on identifying and extracting explicit numerical indicators of impacts for a specific event, such as the number of fatalities, the monetary value of economic losses, or the affected area (e.g., hectares of land burned). Databases in this category often mirror classifications in sources like EM-DAT[13]. This approach requires methods capable of identifying numeric entities and standardizing their units of measurement.